\documentclass[conference]{IEEEtran}
\IEEEoverridecommandlockouts
\usepackage{cite}
\usepackage{amsmath,amssymb,amsfonts}
\usepackage{algorithmic}
\usepackage{graphicx}
\usepackage{textcomp}
\usepackage{xcolor}
\usepackage{url}
\usepackage{array}
\newcolumntype{C}[1]{>{\centering\arraybackslash}m{#1}}
\usepackage{hyperref}
\hypersetup{
    colorlinks=true,
    linkcolor=black,
    filecolor=black,      
    urlcolor=black,
    citecolor=black,
}
\def\BibTeX{{\rm B\kern-.05em{\sc i\kern-.025em b}\kern-.08em
    T\kern-.1667em\lower.7ex\hbox{E}\kern-.125emX}}

\usepackage{fancyhdr}
\begin{document}

\title{\vspace{8pt}EmojiHeroVR: A Study on Facial Expression Recognition under Partial Occlusion from Head-Mounted Displays \\
\thanks{This study was partially funded as project C4T910 by the Authority for Science, Research, Equality and Districts of the Free and Hanseatic City of Hamburg, Germany (BWFGB).}
}

\author{\IEEEauthorblockN{1\textsuperscript{st} Thorben Ortmann}
\IEEEauthorblockA{
\textit{University of the West of Scotland,} \\
\textit{Hamburg University of Applied Sciences}\\
Hamburg, Germany \\
thorben.ortmann@haw-hamburg.de}
\and
\IEEEauthorblockN{2\textsuperscript{nd} Qi Wang}
\IEEEauthorblockA{
\textit{University of the West of Scotland}\\
Paisley, United Kingdom \\
qi.wang@uws.ac.uk}
\and
\IEEEauthorblockN{3\textsuperscript{rd} Larissa Putzar}
\IEEEauthorblockA{\textit{Hamburg University of Applied Sciences}\\
Hamburg, Germany \\
larissa.putzar@haw-hamburg.de}
}

\maketitle
\thispagestyle{fancy}

\begin{abstract}
Emotion recognition promotes the evaluation and enhancement of Virtual Reality (VR) experiences by providing emotional feedback and enabling advanced personalization.
However, facial expressions are rarely used to recognize users' emotions, as Head-Mounted Displays (HMDs) occlude the upper half of the face.
To address this issue, we conducted a study with 37 participants who played our novel affective VR game \textit{EmojiHeroVR}.
The collected database, \textit{EmoHeVRDB (EmojiHeroVR Database)}, includes 3,556 labeled facial images of 1,778 reenacted emotions.
For each labeled image, we also provide 29 additional frames recorded directly before and after the labeled image to facilitate dynamic Facial Expression Recognition (FER).
Additionally, EmoHeVRDB includes data on the activations of 63 facial expressions captured via the Meta Quest Pro VR headset for each frame.
Leveraging our database, we conducted a baseline evaluation on the static FER classification task with six basic emotions and neutral using the EfficientNet-B0 architecture.
The best model achieved an accuracy of 69.84\% on the test set, indicating that FER under HMD occlusion is feasible but significantly more challenging than conventional FER.
\end{abstract}


\section{Introduction}
Over the last decade, Virtual Reality (VR) has become an established technology in various applications and research areas, including therapy, training, education, entertainment, and human behavior \cite{halbigSystematicReviewPhysiological2021, marin-moralesEmotionRecognitionImmersive2020}.
Through Head-Mounted Displays (HMDs), VR experiences deliver high levels of immersion, presence and interaction while simulating almost arbitrary environments \cite{cipressoPresentFutureVirtual2018}.
These capabilities render VR an ideal tool for the reliable elicitation and study of emotions \cite{somarathnaVirtualRealityEmotion2022}.
Vice versa, emotion recognition is valuable for evaluating and enhancing VR experiences.
Emotions affect human perception, decision-making, behavior and overall psychological and physiological state \cite{Picard97}.
Their recognition promotes in-depth analysis of VR experiences and enables systems to react directly to a user's emotional state, further increasing interactivity and personalization. 
Facial expressions are a very natural and potent signal to convey emotions \cite{ekman2006darwin}.
Automatic Facial Expression Recognition (FER) is a well-researched task with a rich history in computer vision and affective computing \cite{canedoFacialExpressionRecognition2019}.
Modern approaches build upon deep learning models, such as Convolutional Neural Networks (CNNs) and transformer architectures \cite{ortmannFacialExpressionRecognition2023a}.
Typically, FER systems categorize images or image sequences into six to eight emotion categories based on Ekman's theory of basic emotions \cite{ekman1979basic} \cite{liDeepFacialExpression2022}.
While more works focus on static FER, processing single images, dynamic FER systems have demonstrated that including temporal features using multiple frames can be beneficial \cite{kimMultiObjectiveBasedSpatioTemporal2019}.

However, FER is rarely applied in current VR research as HMDs occlude the upper face half, severely limiting the capabilities of conventional FER systems \cite{ortmannFacialEmotionRecognition2023}.
While some works have reported promising results on images with artificial HMD occlusion \cite{yongEmotionRecognitionGamers2019, georgescuRecognizingFacialExpressions2019a, houshmandFacialExpressionRecognition2020c, georgescuTeacherStudentTraining2021, georgescuTeacherStudentTrainingTriplet2021a, gotsmanValenceArousalEstimation2021}, these findings still need to be validated with naturally occluded data in actual VR settings.
Our work addresses this gap by conducting a user study with our novel affective VR game, \textit{EmojiHeroVR}.
We record emotional faces under natural HMD occlusion, demonstrate the practical value of FER in VR environments and evaluate the applicability of existing findings.
Moreover, we contribute our collected database \textit{EmoHeVRDB (EmojiHeroVR Database)}, which is suitable for static, dynamic, and multimodal FER, with baseline evaluations on the static FER task.

\section{Related Work}

\begin{table*}[htbp]
  \caption{Comparison of FER databases}
  \label{tab:databases}
  \begin{center}
  \begin{tabular}{|p{0.09\linewidth}|p{0.24\linewidth}|p{0.03\linewidth}|p{0.03\linewidth}|p{0.04\linewidth}|p{0.03\linewidth}|p{0.15\linewidth}|p{0.2\linewidth}|}
    \hline
    \textbf{Database}&\textbf{Samples}&\textbf{Subj.}&\textbf{Env.}&\textbf{Elicit.}&\textbf{Occl.}&\textbf{Expressions}&\textbf{Annotation}\\
    \hline
    \hline
    KDEF,\linebreak 1998 \cite{lundqvist1998karolinska} & 4,900 images from five angles in two sessions & 70 & Lab & P & N & 6 basic expressions + neutral & Acted by subjects\\
    \hline
    CK+,\linebreak 2010 \cite{luceyExtendedCohnKanadeDataset2010} & 593 image sequences (327 with emotion labels) & 123 & Lab & P & N & 6 basic expressions + neutral and contempt & Acted by subjects\\
    \hline
    RaFD,\linebreak 2010 \cite{langnerPresentationValidationRadboud2010} & 8,040 images from five angles with three gaze directions & 67 & Lab & P & N & 6 basic expressions + neutral and contempt & Acted by subjects\\
    \hline
    FER+,\linebreak 2016 \cite{barsoumFERPLUSTrainingDeep2016} & 35,887 images based on keyword search via Google image search & N/A & Web & (P\&)S & V & 6 basic expressions + neutral and contempt & Each image labeled by 10 annotators\\
    \hline   
    RAF-DB,\linebreak 2017 \cite{liReliableCrowdsourcingDeep2017} & 29,672 images based on keyword search via Flickr image search  & N/A & Web & (P\&)S & V & 6 basic + 12 compound expressions + neutral   & Each image labeled by 40 of 315 annotators\\
    \hline
    AffectNet,\linebreak 2019 \cite{mollahosseiniAffectNetDatabaseFacial2019a} & 291,651 manually labeled images based on keyword search via Google, Bing and Yahoo image search & N/A & Web & (P\&)S & V & 6 basic expressions + neutral and contempt + valence and arousal & Each image labeled by one of twelve annotators; 36,000 images labeled by two annotators\\
    \hline
    EmoHeVRDB,\linebreak 2024 & 3,556 image sequences from two angles + 1,727 facial expression activation sequences  & 37 & Lab & P & HMD & 6 basic expressions + neutral& Acted by subjects and each image labeled by three annotators\\
    \hline
    \multicolumn{8}{l}{Subj.=Subjects, Env.=Environment, Elicit.=Elicitation Method, Occl.=Occlusion, N/A=Not Applicable, P=Posed, S=Spontaneous, N=None, V=Various}
  \end{tabular}
  \end{center}
\end{table*}

Few works have been published regarding the particular case of FER under partial occlusion from HMDs.
Only one \cite{granatoEmpiricalStudyPlayers2020a} of the 42 studies included in the comprehensive review on emotion recognition in VR conducted by Mar{\'i}n-Morales et al. applied FER \cite{marin-moralesEmotionRecognitionImmersive2020}.
However, in \cite{granatoEmpiricalStudyPlayers2020a}, Granato et al. did not apply image-based FER but measured facial muscle activity during VR racing games via Electromyography (EMG) to predict arousal and valence values.
In  \cite{ortmannFacialEmotionRecognition2023}, we analyzed 21 studies in our systematic literature review on FER in VR in detail.
Most studies employed sensors attached to or embedded in HMDs, most prominently electrodes for EMG.
A minority of six works \cite{yongEmotionRecognitionGamers2019, georgescuRecognizingFacialExpressions2019a, houshmandFacialExpressionRecognition2020c, georgescuTeacherStudentTraining2021, georgescuTeacherStudentTrainingTriplet2021a, gotsmanValenceArousalEstimation2021} relied on conventional FER based on images recorded by external cameras.
None conducted a user study.
Instead, all works used existing FER image databases to simulate HMD occlusion artificially.
In \cite{georgescuRecognizingFacialExpressions2019a}, Georgescu et al. experimented with the VGG-face network \cite{parkhiDeepFaceRecognition2015a}, a variant of the VGG-16 architecture \cite{Simonyan15} extensively pre-trained for face recognition on the VGG-face dataset.
Using the 8-class AffectNet database \cite{mollahosseiniAffectNetDatabaseFacial2019a}, they first trained and evaluated their model with unoccluded data, achieving an accuracy of 59.03\%.
To simulate HMD occlusion, they blacked out the entire upper half of the images.
When evaluated against the occluded test set, the model's accuracy dropped to 37.70\%.
However, when also trained with occluded data, the model's accuracy only decreased to 49.23\%.
For the FER+ dataset \cite{barsoumFERPLUSTrainingDeep2016}, accuracy even only decreased from 84.79\% to 82.28\%.
Houshamand et al. performed a more sophisticated HMD simulation based on facial landmarks to accurately add a black rectangle to each image \cite{houshmandFacialExpressionRecognition2020c}.
They reported results similar to those of Georgescu et al., with VGG-face and ResNet50 \cite{He_2016_CVPR} architectures on the AffectNet, FER+, and RAF-DB \cite{liReliableCrowdsourcingDeep2017} datasets. 

Large, diverse and well-designed databases are the basis for the development of robust FER models.
Early FER databases, such as KDEF \cite{lundqvist1998karolinska}, JAFFE \cite{lyonsCodingFacialExpressions1998a}, CK \cite{kanadeCKComprehensiveDatabase2000}, CK+ \cite{luceyExtendedCohnKanadeDataset2010}, and RaFD \cite{langnerPresentationValidationRadboud2010}, laid the groundwork by recording posed emotions under lab-controlled conditions.
However, sample sizes were still small for deep learning, real-world conditions like varying angles, illumination, and occlusions were not represented enough, and posed emotions differed from spontaneous ones.
Consequently, web-based and crowd-sourced databases like FER-2013 \cite{goodfellowChallengesRepresentationLearning2013a}, FER+, RAF-DB and AffectNet were constructed.
Researchers acquired large amounts of image data by keyword-based image search using engines like Google, Bing, Yahoo, and Flickr.
One of the largest and most widely used databases is AffectNet.
It comprises about one million facial images.
About 420,000 were manually annotated, resulting in 291,651 images with expression labels for eight categories and arousal and valence values. 
Web-based image databases are generally more diverse and natural than lab-controlled ones.
Facial expressions are more spontaneous, although depending on the search results, web-based databases include posed expressions as well.
While various natural occlusions, such as sunglasses or hands, are present in these databases, no database focuses on HMD occlusion.
Thus, the works concerned with FER under HMD occlusion applied artificial occlusions to simulate HMDs.
Our database, EmoHeVRDB, follows the traditional approach of recording posed expressions under lab-controlled conditions.
However, it is the first to provide naturally HMD-occluded emotional faces.
Table \ref{tab:databases} provides a comprehensive comparison of key FER databases and EmoHeVRDB.

\section{Study Preparation}
We prepared a user study to collect naturally HMD-occluded emotional faces.
We developed the VR game EmojiHeroVR to elicit and record posed emotions, trained an FER model on artificially occluded data and arranged the experimental setup.

\subsection{The Game - EmojiHeroVR}
\begin{figure*}[ht]
  \centerline{\includegraphics[width=\linewidth]{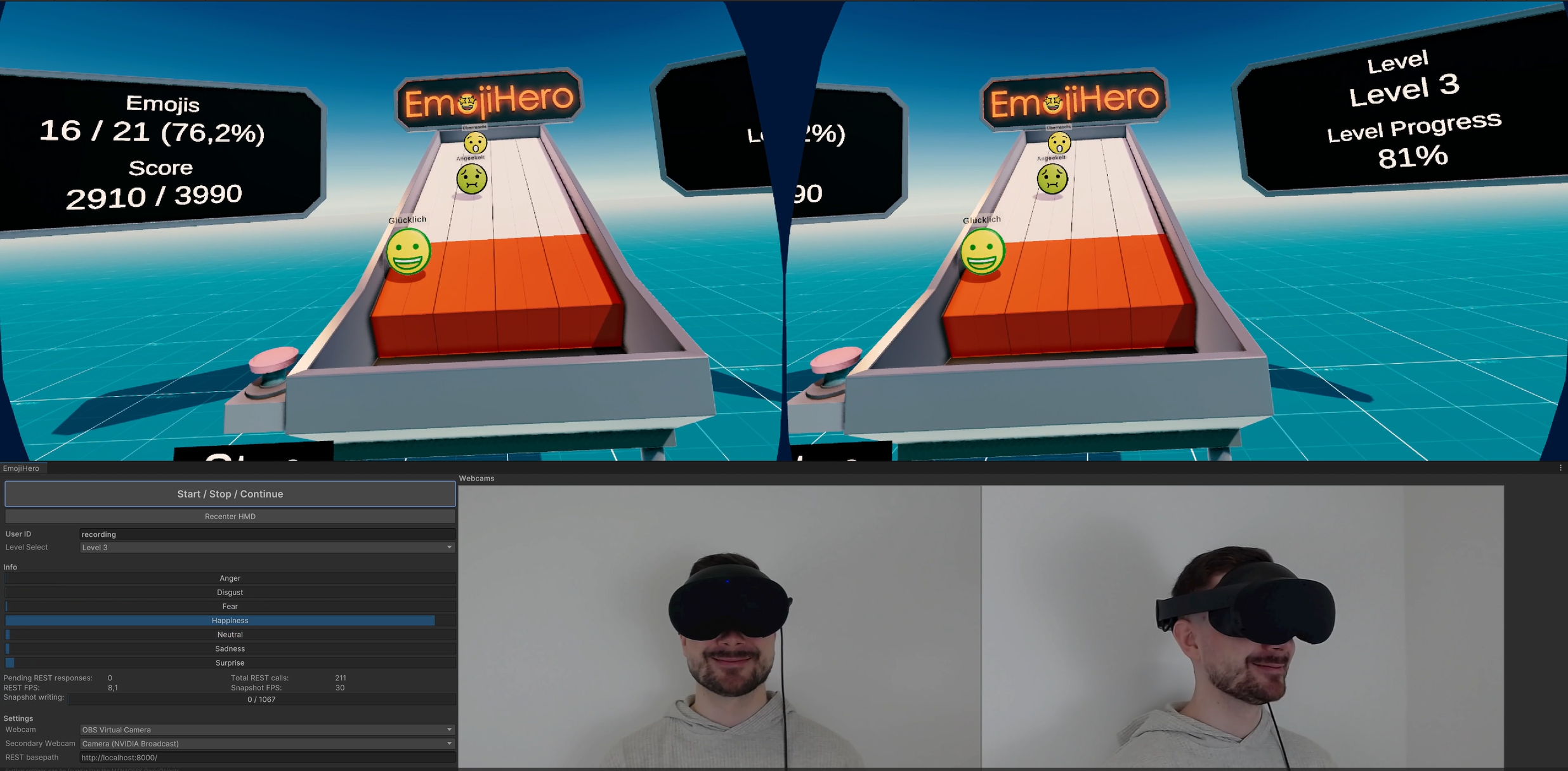}}
  \caption{Captured screen of the Unity Editor during the study conduction - top left: left eye view, top right: right eye view, bottom left: FER results, bottom middle: central recording, bottom right: 45° side-view recording.}
  \label{fig:emoji-hero-gameplay}
\end{figure*}
EmojiHeroVR is a single-player high-score game.
Its name and design are inspired by the popular Guitar Hero video game series.
However, we chose a more neutral design, shown in Fig.~\ref{fig:emoji-hero-gameplay}, with a pinball-machine-like object as the main component.
Emojis spawn in its back and move toward the player in one of four lanes.
When an emoji reaches the red zone directly in front of the player, the player has about one second to reenact the emotion symbolized by the emoji.
The emojis used correspond to the six basic emotions of anger, disgust, fear, happiness, sadness, and surprise, plus neutral, which are commonly used in FER research.
For our study, we designed four levels comprising a predetermined order of emojis.
The levels become progressively more challenging to elevate players' engagement as the number of emojis per level and their movement speed increase while emojis also spawn more frequently.
In total, all four levels comprise 70 emojis or 10 per emotion category.
We developed EmojiHeroVR using the Unity3D game engine.
The FER model, employed to determine whether a reenactment was successful, was developed in Python and made available via a minimalistic web service built with the FastAPI framework.

\subsection{Facial Expression Recognition Model} \label{subsec:fer-model}
\begin{figure}[ht]
  \centering
    \begin{minipage}{.32\linewidth}
    \centering
    \includegraphics[width=\linewidth]{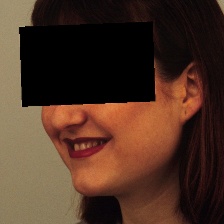}
  \end{minipage}
  \hfill
  \begin{minipage}{.32\linewidth}
    \centering
    \includegraphics[width=\linewidth]{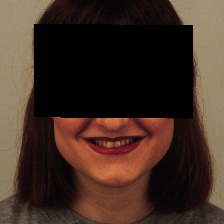}
  \end{minipage}
  \hfill
  \begin{minipage}{.32\linewidth}
    \centering
    \includegraphics[width=\linewidth]{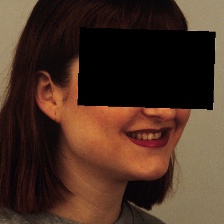}
  \end{minipage}
  \caption{Artificially occluded images from KDEF \cite{lundqvist1998karolinska} (image IDs from left to right: AF10HAHL, AF10HAS, AF10HAHR).
  }
  \label{fig:kdef}
\end{figure}

EmojiHeroVR's game mechanism depends on an FER model to rate whether players reenact emotions successfully.
Following the approach of related works, we prepared a static FER model for HMD occlusion by training with artificially occluded data.
We built upon the Poster++ model architecture \cite{maoPOSTERSimplerStronger2023} due to its state-of-the-art results on RAF-DB and AffectNet, its comparably low computational complexity and code availability.
We combined data from the AffectNet, EmotioNet \cite{Benitez-Quiroz_2016_CVPR}, ExpW \cite{zhangFacialExpressionRecognition2018}, FER+, and SFEW \cite{dhallStaticFacialExpression2011} datasets for the seven emotion categories used in EmojiHeroVR.
Subsequently, we performed face and facial keypoint detection for each image using Google's MediaPipe Python package to crop and occlude images uniformly.
To accurately simulate HMD occlusion, we followed the same approach as \cite{houshmandFacialExpressionRecognition2020c} and \cite{gotsmanValenceArousalEstimation2021} and added a black rectangle to each image based on the facial keypoint detection result.
Consequently, our imbalanced training set comprised about 360,000 images that we used alongside a balanced validation set of 5,600 images to train our model.
The training set was highly imbalanced due to the imbalance of the underlying datasets, especially AffectNet.
We accounted for this by setting each sample's selection probability inversely proportional to its class frequency, in combination with data augmentation techniques, as Mao et al. did in \cite{maoPOSTERSimplerStronger2023}.
We also adjusted the class weights of the Categorical Cross Entropy loss function to favor the anger, fear and sadness classes.
Following this training approach, our model achieved a validation accuracy of 61\%, with no class's F-Score being below 55\%.
To further validate our model, we performed cross-dataset evaluation against HMD-occluded versions of the 2,940 central and 45° side-view images of the KDEF dataset, exemplarily shown in Fig.~\ref{fig:kdef}.
We chose KDEF because it is balanced, and its recording conditions are similar to those of our study.
Our model achieved a cross-dataset evaluation accuracy of 55\%.
After resuming the training process with a very low learning rate and the occluded images of 62 out of 70 subjects of the KDEF dataset, the model's accuracy on the remaining occluded KDEF data rose to 81\%, providing us with confidence to apply the model in EmojiHeroVR.

\subsection{Experimental Setup}
\begin{figure}[ht]
  \centering
  \includegraphics[width=\linewidth]{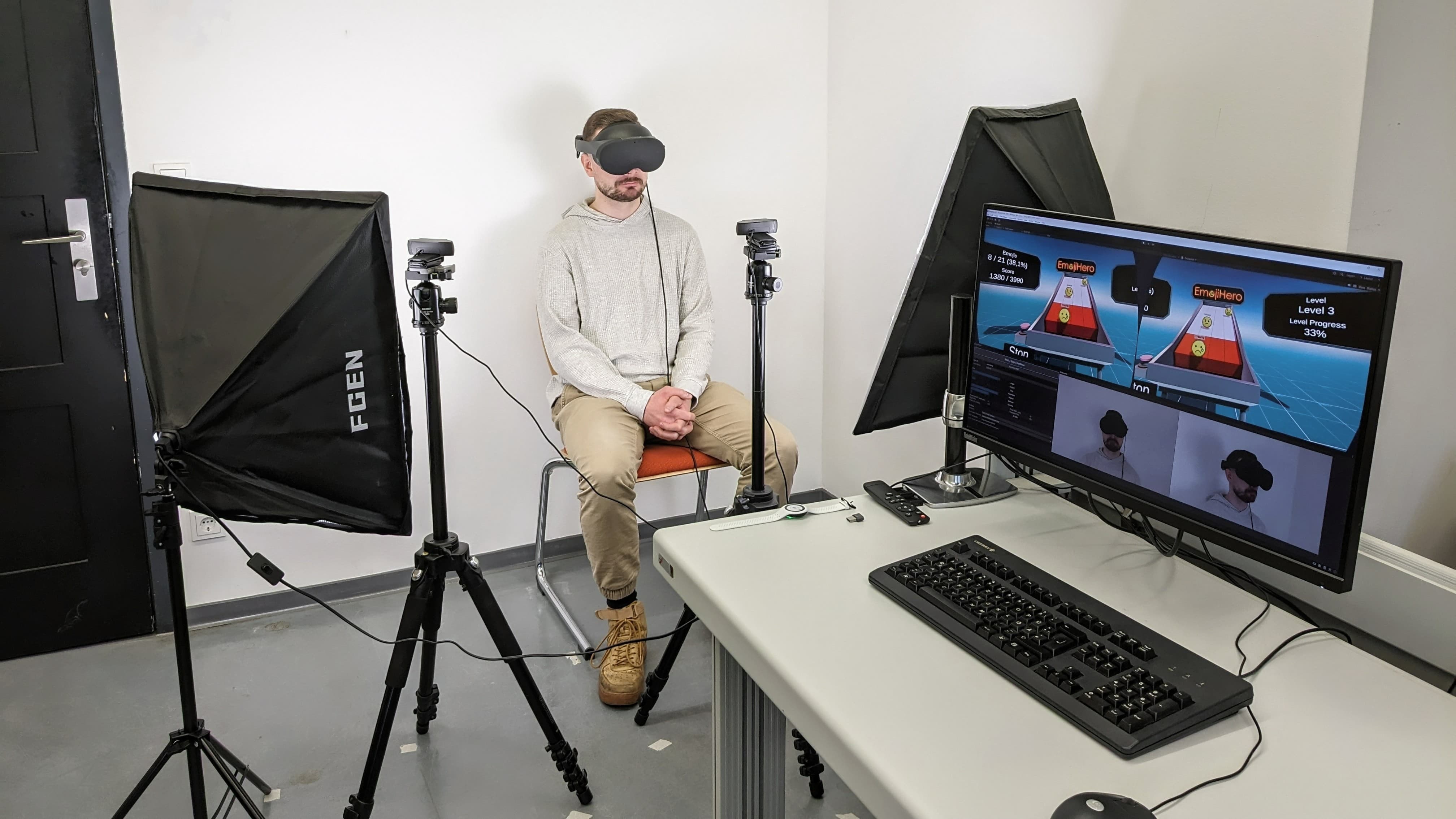}
  \caption{Experimental setup with two cameras, two softboxes, a PC and a participant wearing the Meta Quest Pro VR headset.}
  \label{fig:tech-set}
\end{figure}
We used two consumer Logitech webcams (models C930e and C920e) to record participants' faces from a frontal and a 45° side view at 30 frames per second (FPS) and a resolution of 1280x720 pixels.
Two softboxes ensured good illumination at about 5400K color temperature.
We employed the Meta Quest Pro as the VR headset for our study because it can capture facial expressions by accessing its embedded sensors via the Meta XR Core SDK's Face Tracking API \cite{meta_face_tracking}.
As depicted in Fig.~\ref{fig:tech-set}, both cameras and the headset were connected to a PC, which ran EmojiHeroVR as a Unity application and the Python web service providing the FER model.
The PC had an Intel i9-12900K CPU, 64GB of DDR4 RAM, and an Nvidia GeForce RTX 3090 GPU and ran Windows 11 version 22H2.

\subsection{Sample Selection}
We primarily recruited participants in the environment of the Hamburg University of Applied Sciences (HAW Hamburg).
We promoted the study via e-mail using mailing lists, in lectures, and by prominently placing posters and flyers around the HAW Hamburg's Finkenau campus.
Participants were compensated with €10 in cash to value their contribution.
The recruitment process, the study's conduction, and the corresponding data processing were approved by the ethics committees of the University of the West of Scotland (application 21639) and of the HAW Hamburg (application 2023-25).

\section{Study Execution and Results}
We conducted the study on eight days between the 15\textsuperscript{th} of November and the 1\textsuperscript{st} of December 2023 in the facilities of the HAW Hamburg's Finkenau campus.
Thirty-seven subjects, primarily students and faculty members, participated.
Twenty participants reported being male, fifteen reported being female, and two chose not to disclose their gender.
The age ranged from 19 to 50 years ($\mu=27.19$ and $\sigma=7.66$).
Most subjects identified with a European heritage.

\subsection{Procedure}
First, participants were briefed on the study's procedure and objectives.
Next, the research team handed out participant information sheets and consent forms detailing the study design, particularly emphasizing data processing practices.
After giving informed consent, participants completed a short questionnaire focusing on demographic data.
Then, they were introduced to the Facial Action Coding System (FACS) \cite{1370848662448368390} and trained on accurately reenacting the seven emotions featured in EmojiHeroVR for about 10 minutes.
Subsequently, the research team positioned participants at the recording station, adjusted the cameras, and set up the VR headset.
After a brief trial session, participants played through four levels, taking about 90-second breaks between each level.
The duration of levels increased from level one, with about 25 seconds and nine emojis, to level four, with about 50 seconds and 28 emojis.
Upon completion, participants answered the Game Experience Questionnaire (GEQ) \cite{ijsselsteijn2013game} and the Virtual Reality Sickness Questionnaire (VRSQ) \cite{kim2018virtual} to assess their game experience and any feelings of VR sickness.

\subsection{Collected Data}
Both cameras recorded participants' faces at 30 FPS with a resolution of 1280x720 pixels during their gameplay.
We directly associated every recorded frame with its corresponding emoji or reenactment process by defining a recording window for each emoji.
An emoji's recording window equals the time the emoji is present in the red reenactment zone displayed in the game's graphical user interface, plus about half a second upfront.
We saved every frame inside an emojis recording window in a corresponding directory as a PNG file with a UNIX timestamp plus a camera index as its name.
Each frame outside a recording window was discarded.
For level one, the average length of the recording window is 2.13 seconds,
resulting in 64.63 frames ($\sigma=0.61$) on average per camera.
For level four, the average length of the recording window is 1.59 seconds,
resulting in 48.52 frames ($\sigma=0.63$) on average per camera.
The differences are due to increasing movement speed and decreasing spawn rate of emojis between levels one and four.
In total, we recorded 37 participants reenacting 2,590 emojis, resulting in 147,490 images per camera.
In addition to recording with external cameras, we also captured facial expressions using the Meta Quest Pro's sensors.
The Meta XR Core SDK's Face Tracking API was accessed in each executed update loop of the game inside an emoji's recording window, averaging 62.78 times per second.
In consequence, a total of 332,734 API calls were executed during the study conduction.
Due to a technical malfunction, no face-tracking data were captured for the seventh participant.  
For each API call, we saved a list of 63 floating-point numbers ranging from 0 to 1.
Each number represents the measured activation strength of one of 63 facial expressions based on FACS.
Typically, these measured facial expression activations are utilized to adjust blend shapes, allowing for a detailed and accurate mapping of a user's facial movements onto a 3D model.

\section{Data Annotation}
Our collected data have several limitations.
First, emotions are not spontaneous but posed by participants who are not trained actors.
Authentically reenacting emotions is a challenging task, especially under time pressure caused by the game mechanism of EmojiHeroVR.
Second, the challenge during the gameplay might elicit spontaneous emotions contrary to the emotion to pose, for example, stress or frustration when a player fails to reenact an emoji correctly.
Third, players might quickly learn how to please the game or the underlying FER model to get a higher score and adjust their original facial expressions accordingly, resulting in data biased towards the employed FER model.
To compensate for these limitations and ensure high data quality, three student annotators additionally labeled the collected data.

\subsection{Procedure}
We decided to label one central-view image per reenacted emoji.
To select the image to label, we used our FER model, described in subsection~\ref{subsec:fer-model}, to predict a 7-dimensional stochastic vector for each image.
Then, per emoji, we chose the image with the highest probability for the emotion to be reenacted.
We excluded the first 15 frames per emoji from the selection as those were recorded before an emoji entered the red reenactment zone in the game.
Also, this ensures that at least 15 frames are available as context before each labeled image.
Each of the 2,590 selected images was labeled by three students with experience in the field of FER.
For labeling, we developed a web service using the Python Flask framework that draws images in random order, shows the image to label, and lets the user label the image by clicking on one of seven buttons labeled with the corresponding emotion category.

\subsection{Results}
On average, annotators recognized the emotion to be reenacted in 71.87\% of cases.
\begin{figure}[ht]
  \centering
  \includegraphics[width=\linewidth]{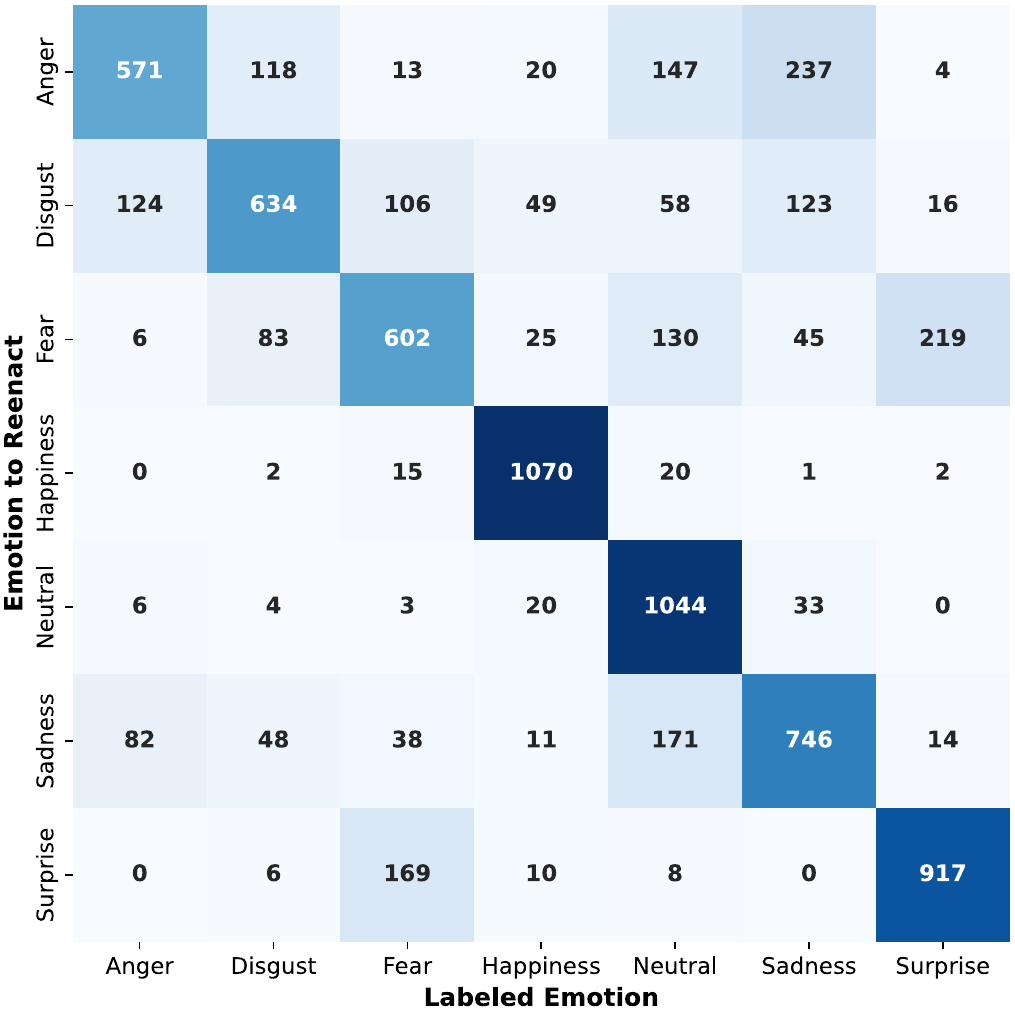}
  \caption{Confusion matrix - emotion to reenact vs labeled emotion.}
  \label{fig:label-cm}
\end{figure}
As visualized in Fig. \ref{fig:label-cm}, the most recognizable emotions were happiness, neutral, and surprise, with F-scores of 92.44\%, 77.68\% and 80.37\%, respectively.
In contrast, with recall values between 51.44\% and 57.12\%, anger, fear and disgust were often not recognized.
However, Cohen's Kappa scores for annotator pairs, listed in \mbox{Table \ref{tab:cohen}}, and a Fleiss' Kappa score of 67.76\% indicate a substantial agreement and provide confidence in the given labels.
\begin{table}[ht]
\setlength{\tabcolsep}{10pt}
\centering
\caption{Cohen's Kappa Scores}
\label{tab:cohen}
\begin{tabular}{ccc}
\hline
\textbf{Annotator} & \textbf{Annotator} & \textbf{Score} \\ \hline
1 & 2 & 67.32\% \\
1 & 3 & 63.33\% \\
2 & 3 & 72.82\% \\ \hline
\multicolumn{2}{c}{Average} & 67.82\% \\ \hline
\end{tabular}
\end{table}

\section{Database Construction}
To ensure high data quality in our database, we discarded all samples for which not at least two of three annotators agreed to recognize the emotion that was to be reenacted.
As a result, 1,921 labeled images with a Fleiss' Kappa Score of 78.60\% remained.
Due to this selection step, the data are unbalanced in favor of emotions that were easier to reenact and recognize.

\subsection{Training, Validation and Test Split}
\begin{figure}[ht]
  \centering
  \includegraphics[width=\linewidth]{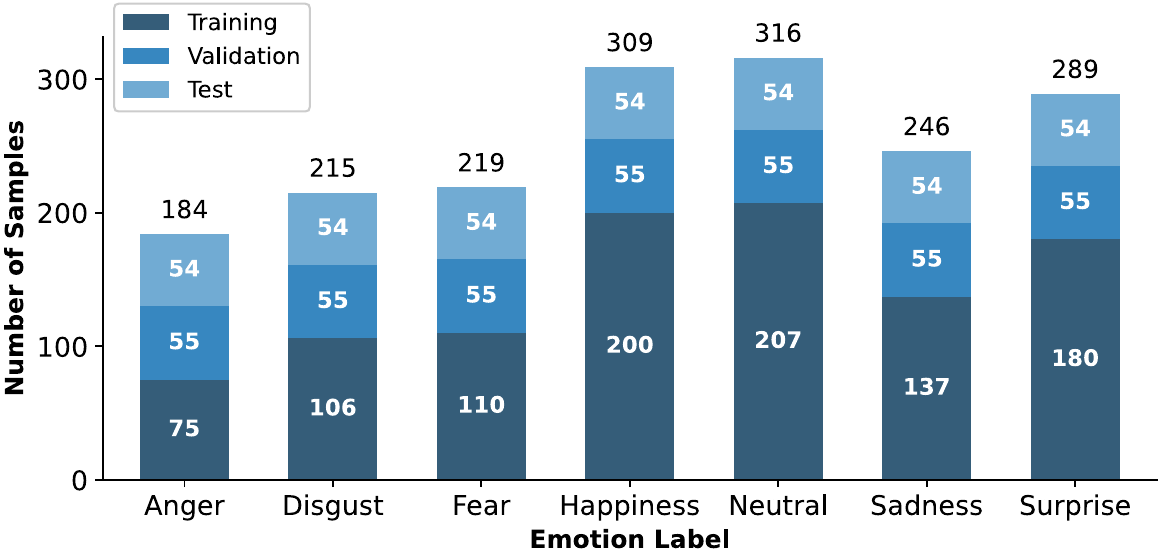}
  \caption{Label distribution for the training, validation and test sets.}
  \label{fig:label-dis}
\end{figure}

We split our data into a training, a validation, and a test set to establish a comparable baseline.
To prevent information leakage, we divided the 37 participants into a training set of 21, a validation set of 8 and a test set of 8.
We considered the available age, gender, and ethnicity data for a stratified split and aimed for the validation and test sets to be as balanced as possible.
Subsequently, to balance the validation and test sets, we iteratively discarded random samples of the most represented participant for each class until the number of samples was equal to the lowest class frequency.
Consequently, we discarded 64 of 449 samples from the validation set and 79 of 457 samples from the test set.
As a result, EmoHeVRDB comprises an unbalanced training set of 1015 labels, a balanced validation set of 385 labels and a balanced test set of 378 labels.
The final label distribution is visualized in Fig. \ref{fig:label-dis}.

\subsection{Construction and Statistics}
Based on the 1,778 labels for central-view images, we constructed the rest of EmoHeVRDB.
First, we added the side-view recordings with identical timestamps as the labeled images, doubling the number of samples to 3,556.
Applying face detection with the MediaPipe Python package, we cropped all images to 720x720 pixels centered at the face.
Subsequently, we converted the PNG files to JPG with a compression quality of 95\% to reduce file size.
Next, we prepared our database for dynamic FER.
For each labeled image, we selected 29 contextual frames from the corresponding recording window to cover about one second of facial movements for each reenactment.
We added as many preceding frames as possible and increased the count to 30 with subsequent frames if necessary.
Consequently, EmoHeVRDB includes 3,556 30-frame sequences recorded from two perspectives.
Lastly, we added the data captured via the Meta Quest Pro to EmoHeVRDB.
Using the recorded timestamps, we associated a list of 63 facial expression activations with each included frame.
Notably, this was not possible for recordings of the seventh participant, as no face-tracking data were captured due to a technical malfunction.
In total, we provide 1,727 30-element facial expression activation sequences as JSON files.

\section{Baseline}
We performed several experiments with EmoHeVRDB to create a baseline for the image-based static FER task and examine the effects of natural HMD occlusion.
For comparability, reproducibility, and fast training, we used the EfficientNet-B0 architecture for all experiments.
The EfficientNets model family has achieved state-of-the-art results on several computer vision tasks while being highly efficient \cite{tan2019efficientnet}.
B0 is its smallest variant.
In \cite{savchenkoFacialExpressionAttributes2021}, Savchenko reported an accuracy of  61.32\% for the 8-class FER task on AffectNet using an EfficientNet-B0 ($\sim$5.3M parameters) pre-trained on the VGGFace2 dataset \cite{Cao18}.
Thereby, it outperforms the accuracy of 59.03\% reported by \cite{georgescuRecognizingFacialExpressions2019a} using the much larger VGG-face model ($\sim$138.4M parameters).
We implemented our experiments with TensorFlow 2.15, leveraging Keras' standard implementation of EfficientNet-B0.
All experiments were executed on a machine with an Intel i9-12900K CPU, 64GB of DDR4 RAM, and an Nvidia GeForce RTX 3090 GPU.
It ran Ubuntu 22.04 and had CUDA 12.2 and cuDNN 8.9 installed.

\subsection{Database Preparation}\label{subsec:db-prep}
We prepared five datasets for our experiments.
For face and facial keypoint detection, we used the MediaPipe Python package, version 0.10.
For resizing images with Lanczos interpolation, we used the opencv-python package, version 4.9. 

\subsubsection{AffectNet-7}
We built the 7-class AffectNet dataset by removing all images of the contempt class from the 8-class AffectNet dataset, resulting in 287,401 available images.
As typically done, we used the predefined validation set, comprising 500 images per class, as the test set.
We randomly split 380 samples, which is 10\% of the lowest class frequency, from each class to create a balanced validation set.
All images have a resolution of 224x224 pixels.

\subsubsection{AffectNet-7-occl}
We created an artificially occluded version of AffectNet-7 by adding a black rectangle to each image based on face and facial keypoint detection results, similar to the approaches of \cite{houshmandFacialExpressionRecognition2020c} and \cite{gotsmanValenceArousalEstimation2021}.
If no face was detected or the detection results were not plausible, we defaulted to occluding the top 54\% of the image at 90\% of its width.

\subsubsection{KDEF-SHR}
The original KDEF dataset contains 4,900 facial images of 70 subjects for seven emotion categories recorded from five angles.
For our experiments, KDEF served as a reference for the results on EmoHeVRDB.
So, we prepared it to resemble EmoHeVRDB as closely as possible.
Thus, we selected only the 1,960 central and 45° left-side-view images, coded with \textit{S} for \textit{straight} and \textit{HR} for \textit{half right profile}, respectively, for our experiments.
We cropped the images closer to the face and resized them to 224x224 pixels to make them uniform with the AffectNet images.
Subsequently, we split the data into training, validation, and test sets by randomly selecting 14 out of 70 subjects, seven female and seven male, for the validation and the test set each.

\subsubsection{KDEF-SHR-occl}
Similar to AffectNet-7-occl, we also created an occluded version of our prepared KDEF-SHR dataset using the same occlusion approach.

\subsubsection{EmoHeVRDB}
We selected the 3,556 labeled central and 45° side-view images from EmoHeVRDB, cropped them closer to the face and downsized them to 224x224 pixels.

\subsection{Training}
Similar to \cite{georgescuRecognizingFacialExpressions2019a, houshmandFacialExpressionRecognition2020c, gotsmanValenceArousalEstimation2021, savchenkoFacialExpressionAttributes2021}, we applied a transfer learning approach in our experiments.
We initialized an EfficientNet-B0, pre-trained on the ImageNet database \cite{ILSVRC15}, froze its weights and replaced the final Dense layer with a new one with 7 units and softmax activation.
Subsequently, we started fitting the network's top for FER by training for a few epochs with a high learning rate.
Next, we incrementally unfroze more parts of the network to fine-tune it with a lower learning rate.
For all experiments, we used the SparseCategoricalCrossentropy loss function, an Adam optimizer with a learning rate between 1e-3 and 1e-6 and a batch size of 32.
We weighted the loss function with the normalized inverse class frequencies to account for imbalances in the training sets.
Additionally, we incorporated several image augmentation layers, namely RandomFlip, RandomTranslation, RandomRotation, RandomZoom, RandomContrast and RandomBrightness, in our network to increase the variety of the training data.
Generally, we kept the hyperparameter space small and focused on optimizing the fine-tuning process by tuning the combination of layers to unfreeze, epochs to train and the learning rate.

\subsection{Results}
For the first set of experiments, we fine-tuned an EfficientNet-B0 for each of the five datasets listed in subsection \ref{subsec:db-prep} following the training process described in the previous subsection.
Our results on AffectNet, listed in Table \ref{tab:transfer-acc}, demonstrate the effectiveness of our approach.
\begin{table}[ht]
\centering
\setlength{\tabcolsep}{5pt}
\caption{Classification Accuracies on AffectNet}
\label{tab:transfer-acc}
\begin{tabular}{lccccc}
\hline
\textbf{Model} & \textbf{Pre-trained} & \textbf{Occlusion} & \multicolumn{2}{c}{\textbf{Accuracy, \%}} \\
& & & \textbf{7 classes} & \textbf{8 classes} \\ \hline
EfficientNet-B0 \cite{savchenko2022classifying} & ImageNet & No & 60.80 & 57.55 \\
VGG-face \cite{georgescuRecognizingFacialExpressions2019a} & VGGFace & No & - & 59.03 \\
VGG-face \cite{houshmandFacialExpressionRecognition2020c} & VGGFace & Yes & - & 50.13 \\
\hline
Our EfficientNet-B0 & ImageNet & No & 64.31 & - \\ 
Our EfficientNet-B0 & ImageNet & Yes & 52.57 & - \\ 
\hline
\end{tabular}
\end{table}
We achieve higher accuracies than related works on unoccluded and occluded versions of AffectNet.
However, in contrast to us, \cite{savchenko2022classifying, georgescuRecognizingFacialExpressions2019a, houshmandFacialExpressionRecognition2020c} operated on the 8-class AffectNet dataset.
Similar to \cite{georgescuRecognizingFacialExpressions2019a}, we find an accuracy decrease of about 10\% when working with the occluded version of AffectNet. 

FER on KDEF is significantly easier.
Good recording conditions and very expressive faces allow our model to achieve 86.48\% accuracy on the test set of KDEF-SHR.
The accuracy only decreases to 79.59\% on KDEF-SHR-occl.
\begin{table}[ht]
\centering
\setlength{\tabcolsep}{5pt}
\caption{Classification Accuracies on KDEF and EmoHeVRDB}
\label{tab:transfer-acc-kdef}
\begin{tabular}{lccc}
\hline
\textbf{Model} & \textbf{Pre-trained} & \textbf{Dataset} & \textbf{Accuracy, \%} \\
\hline
Our EfficientNet-B0 & ImageNet & KDEF-SHR & 86.48 \\
Our EfficientNet-B0 & ImageNet & KDEF-SHR-occl & 79.59 \\
Our EfficientNet-B0 & ImageNet & EmoHeVRDB & 69.84 \\
\hline
\end{tabular}
\end{table}
We assume that AffectNet-7 is more affected by occlusion than KDEF-SHR because its images contain fewer distinctive features.
Thus, additional occlusions are more likely to hide the only crucial information available.

Our baseline accuracy on EmoHeVRDB is 69.84\%.
Our model performs worst in the anger class, with an F-score of 49.11\%, followed by the disgust class, with 61.67\%.
The surprise, happiness, and neutral classes are the most recognizable, with F-Scores of 86.12\%, 77.18\%, and 76.42\%, respectively.
The overall accuracy is 16.64\% lower than on KDEF-SHR, which was recorded under similar conditions, indicating that FER under natural HMD occlusion is a substantially harder task than regular FER.
However, the accuracy on EmoHeVRDB is also 9.75\% lower than on KDEF-SHR-occl, which we constructed to resemble EmoHeVRDB.
This difference is likely due to the higher data variety in EmoHeVRDB.
The subjects recorded for KDEF were trained amateur actors instructed to pose strong and clear facial expressions.
They were between 20 and 30 years old and wore no beards, mustaches, earrings, eyeglasses, or visible makeup \cite{goeleven2008}.
In contrast, for EmoHeVRDB, the age ranged from 19 to 50, and subjects wore various beards, makeup, earrings, and similar jewelry.
Additionally, emotions were posed under time pressure while playing a VR game, resulting in a larger variety of facial expressions and head poses.      

For the second set of experiments, we performed cross-dataset evaluations with the models trained on AffectNet-7 and AffectNet-7-occl against KDEF-SHR, KDEF-SHR-occl and EmoHeVRDB.
\begin{table}[ht]
\centering
\setlength{\tabcolsep}{5pt}
\caption{Cross-Dataset Evaluations}
\label{tab:cross-acc-kdef}
\begin{tabular}{m{2.15cm}ccccc}
\hline
\textbf{Trained} & \textbf{Evaluated} & \multicolumn{4}{c}{\textbf{Accuracy, \%}} \\
& & \textbf{Train} & \textbf{Val} & \textbf{Test} & \textbf{Avg} \\
\hline
Affectnet-7 & KDEF-SHR & 80.68 & 83.38 & 83.67 & 81.82 \\
Affectnet-7 & KDEF-SHR-occl & 52.68 & 59.08 & 58.16 & 55.06 \\
Affectnet-7-occl & KDEF-SHR-occl & 62.72 & 67.26 & 65.05 & 64.10 \\
Affectnet-7 & EmoHeVRDB & - & 33.38 & 37.57 & 35.45 \\
Affectnet-7-occl & EmoHeVRDB & - & 32.60 & 36.90 & 34.73 \\
\hline
Affectnet-7-occl \& \newline EmoHeVRDB & EmoHeVRDB & - & - & 74.74 & 74.74 \\
\hline
\end{tabular}
\end{table}
Applying the AffectNet-7 model on KDEF-SHR works very well, resulting in an accuracy of 81.82\%.
The AffectNet-7-occl model's accuracy on KDEF-SHR-occl is significantly lower, but the model still classifies 64.10\% of the samples correctly.
Most notably, the AffectNet-7 and AffectNet-7-occl models, perform equally poorly on EmoHeVRDB, with accuracies of about 35\%.
This implies a significant difference in the data of AffectNet-7(-occl) and EmoHeVRDB.
To investigate this difference, we experimented with training on a combination of AffectNet-7-occl and EmoHeVRDB.
While generally sampling according to the class frequencies of EmoHeVRDB, we varied the ratio between AffectNet-7-occl and EmoHeVRDB samples.
A ratio of 1:1 resulted in 70.11\% accuracy on the test set of EmoHeVRDB.
The best model, trained with a ratio of 1:4, achieves 74.74\% accuracy, outperforming the model trained exclusively on EmoHeVRDB by almost 5\%.
This suggests that the emotion representations in AffectNet-7-occl and EmoHeVRDB do not differ significantly.
On the contrary, the diverse samples from AffectNet-7-occl can enrich and regularize the learning process, helping models to better generalize on EmoHeVRDB.
We conclude that the primary reason for the very low cross-dataset evaluation accuracy is the absence of naturally HMD-occluded faces in AffectNet-7-occl.
It seems that our artificial occlusion approach does not simulate HMDs realistically enough to make models robust against natural HMD occlusion.
These findings underline the importance of EmoHeVRDB for developing reliable FER models for real VR scenarios.

\section{Conclusion and Future Work}
Our work presents the novel database EmoHeVRDB.
It comprises 3,556 image sequences of the emotional faces of 37 participants playing our novel affective VR game, EmojiHeroVR.
Additionally, EmoHeVRDB includes 1,727 facial expression activation sequences captured via the Meta Quest Pro VR headset.
The results of our baseline evaluation on the static FER task demonstrate the success of our study design and subsequent data annotation process in constructing a novel, high-quality database.
Our code and details on how to request access to EmoHeVRDB are available on GitHub:\linebreak \textit{\url{https://github.com/thorbenortmann/emoji-hero-vr-database}}.
A demo video showing the third level of EmojiHeroVR, used in the user study, is available on YouTube: \textit{\url{https://youtu.be/TnJrGYOjKJc}}.
In future work, we plan to leverage the facial expression activations included in EmoHeVRDB, first for static unimodal FER and, subsequently, in combination with the image data for multimodal FER.
Finally, we plan to make use of EmoHeVRDB's sequential character to also experiment with dynamic FER in VR environments.

\section*{Ethical Impact Statement}
The overarching goal of our research is to enhance VR experiences through automatic emotion recognition. Particularly in application areas, such as therapy, education and entertainment, recognizing the user's emotional state can be highly beneficial. Our study contributes to this goal by exploring the feasibility of FER utilizing data collected via an affective VR game.

\subsection{Conducting a User Study}
Our research design included collecting naturally HMD-occluded faces through a user study.
The study design was approved by the ethics committees of the University of the West of Scotland (application 21639) and of the HAW Hamburg (application 2023-25).

\subsubsection{Data Privacy and Informed Consent}
All participants provided informed consent for their data to be used within the context of our research and for their pseudonymized data to be shared with other researchers under controlled conditions. We require all researchers to apply for access to EmoHeVRDB, stating their name, affiliation and intended usage. Access is only granted after confirming an applicant's identity and receiving a signed licensing agreement, which especially does not permit any redistribution of the data without our prior written approval.

\subsubsection{Data Diversity and Bias}
While we did not target any specific group, our study lacks ethnic diversity.
The vast majority of participants identified with a European heritage.
However, we believe this is not due to a biased recruitment process but mirrors our university's environment in Hamburg, Germany.
For the same reason, our sample's average age is significantly lower than the general population's.
Concerning gender, our sample represents slightly more male than female participants.
Consequently, our sample of 37 participants in EmoHeVRDB is not representative of the general population and has to be used with care.
Algorithmic bias is probable when only learning from our data and applying learned models to a broader demographic group.

\subsection{Emotion Modelling and Elicitation}

\subsubsection{Categorical Basic Emotions}
Our study employed a categorical emotion model of six basic emotions plus neutral, which is widely utilized in FER research. Using it enabled us to rely on existing datasets and compare our results with related works. However, we acknowledge that this model carries the risk of oversimplification and may not encompass the cultural and individual variability in emotional expression. 

\subsubsection{Posed Emotions}
In our study, participants posed emotions to gain points in the EmojiHeroVR game. However, authentically reenacting emotions is challenging. Thus, posed emotions can differ significantly from spontaneous ones. To mitigate this and to improve data consistency, participants were introduced to FACS and trained on accurately reenacting emotions for about 10 minutes before playing EmojiHeroVR. Also, the data annotation process provided an additional level to secure data quality. Nevertheless, there remains a gap between the posed emotions collected in our study and real spontaneous emotions.

\subsection{Data Annotation}
To ensure high data quality, three students who have worked in the field of FER before labeled 2,590 images of the collected data to construct the final database. Our university employed all annotators as student or research assistants. While annotators were familiar with various emotion models and FACS, a personal labeling bias is possible. Also, all annotators had a similar cultural background and age. For privacy reasons, their names are not disclosed. To mitigate personal bias, we computed several metrics to ensure substantial labeling agreement.

\subsection{Conclusion}
We acknowledge several limitations of our study, particularly regarding the demographic diversity of the collected data, the difference between posed and spontaneous emotions and the capability of the categorical emotion model to capture all facets of emotions accurately. However, we emphasize our study's role as the first user study to investigate image-based FER under natural HMD occlusion to gain a preliminary understanding of the feasibility of FER in VR settings. We expect a positive impact from reliable automatic emotion recognition in VR as many beneficial applications such as therapy and education may profit from it.
Nevertheless, further developments must carefully consider the capability of highly immersive and affective VR environments to influence people's emotions and possibly opinions.

\bibliographystyle{IEEEtran}

\end{document}